\pdfoutput=1

\documentclass[10pt,twocolumn,letterpaper]{article}

\usepackage[pagenumbers]{cvpr} 

\usepackage{booktabs}
\usepackage{multirow}
\usepackage{float} 
\usepackage{subcaption}

\usepackage[dvipsnames]{xcolor}

\definecolor{cvprblue}{rgb}{0.21,0.49,0.74}
\usepackage[pagebackref,breaklinks,colorlinks,citecolor=cvprblue]{hyperref}

\def\paperID{5337} 
\def\confName{CVPR}
\def\confYear{2024}

\title{The Devil is in the Details: Boosting Guided Depth Super-Resolution via Rethinking Cross-Modal Alignment and Aggregation}

\author{Xinni Jiang$^{1,}$\setcounter{footnote}{1}\thanks{Equal contribution.}\;\quad
Zengsheng Kuang$^{2,}$\footnotemark[2]\;\quad
Chunle Guo$^{1,}$\footnotemark[1]\;\quad \\
Ruixun Zhang$^3$\;\quad
Lei Cai$^4$\;\quad
Xiao Fan$^5$\;\quad
Chongyi Li$^{1,}$\setcounter{footnote}{0}\thanks{Co-corresponding author.}\;
\\
$^1$Nankai University\;\quad
$^2$South China University of Technology\;\quad \\
$^3$Peking University\;\quad
$^4$Henan Institute of Science and Technology\;\quad \\
$^5$MicroBT Electronics Technology Co., Ltd\;
\\
{\tt\small xinnijiang@mail.nankai.edu.cn, 
ftkuangzs@mail.scut.edu.cn, 
zhangruixun@pku.edu.cn,}\\
{\tt\small cailei2014@126.com, 
fanxiao@microbt.com,
\{guochunle, lichongyi\}@nankai.edu.cn}
}

\begin{document}
\maketitle
\begin{abstract}

Guided depth super-resolution (GDSR) involves restoring missing depth details using the high-resolution RGB image of the same scene. 
Previous approaches have struggled with the heterogeneity and complementarity of the multi-modal inputs, and neglected the issues of modal misalignment, geometrical misalignment, and feature selection. 
In this study, we rethink some essential components in GDSR networks and propose a simple yet effective Dynamic Dual Alignment and Aggregation network (D2A2).
D2A2 mainly consists of 
1) a dynamic dual alignment module that adapts to alleviate the modal misalignment via a learnable domain alignment block
and geometrically align cross-modal features by learning the offset;
and 2) a mask-to-pixel feature aggregate module that uses the gated mechanism and pixel attention to filter out irrelevant texture noise from RGB features and combine the useful features with depth features. 
By combining the strengths of RGB and depth features while minimizing disturbance introduced by the RGB image, our method with simple reuse and redesign of basic components achieves state-of-the-art performance on multiple benchmark datasets.
The code is available at \url{https://github.com/JiangXinni/D2A2}.

\end{abstract}

\section{Introduction}
\label{sec:intro}

Depth maps are widely used in computer vision and computer graphics such as autonomous driving~\cite{godard2017unsupervised, zhang2023lite}, 3D reconstruction~\cite{zheng2023hairstep, ju2023dgrecon, min2022traffic}, 
semantic segmentation~\cite{gupta2014learning, zhang2023cmx, zhou2022canet, chen2020bi}, and scene understanding~\cite{zhou2023bcinet, song2017semantic}, owing to  their spatial geometric information. 
However, the depth maps acquired by consumer-grade depth sensors, such as Microsoft's Kinect and Time of Flight (ToF), are typically low-resolution (LR) and noisy, rendering them unsuitable for vision tasks.
Unlike depth sensors, it is relatively easy to acquire HR RGB images at a low cost. 
In addition, RGB images and their corresponding depth maps have strong structural similarities.
Therefore, RGB images are frequently used to guide the reconstruction of depth maps, which is known as guided depth super-resolution (GDSR).
GDSR has garnered considerable attention from  academia and industry as a cross-modal challenging ill-posed problem.

\begin{figure}
\begin{center}

    \subcaptionbox{RGB / Depth GT}{
		\includegraphics[width=0.48\linewidth]{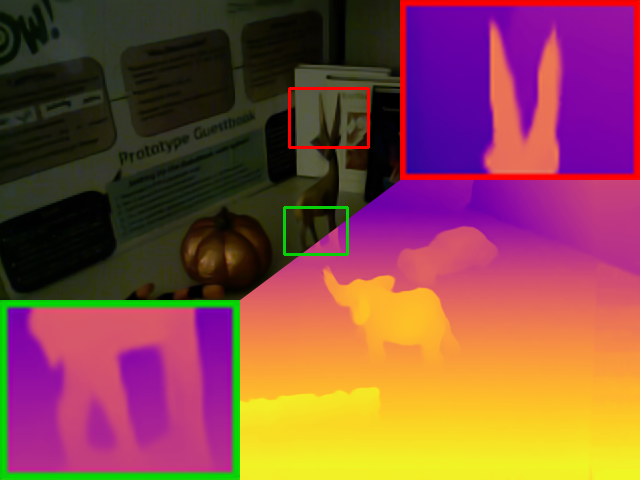}}
    \hfill
    \subcaptionbox{DCTNet (CVPR'22)~\cite{DCTNet}}{
		\includegraphics[width=0.48\linewidth]{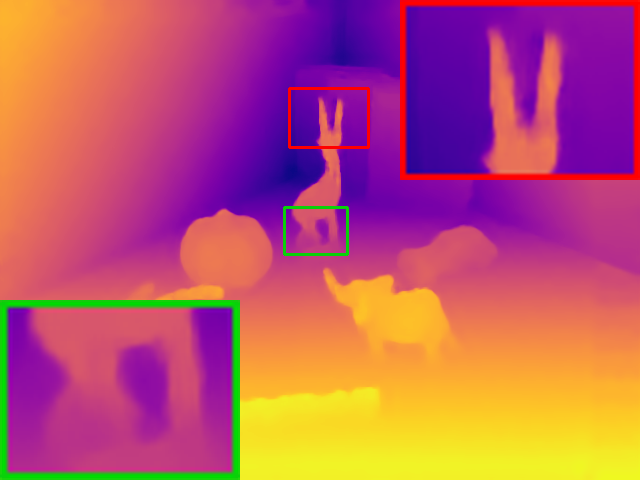}}
    \\
    \subcaptionbox{DADA (CVPR'23)~\cite{DADA}}{
		\includegraphics[width=0.48\linewidth]{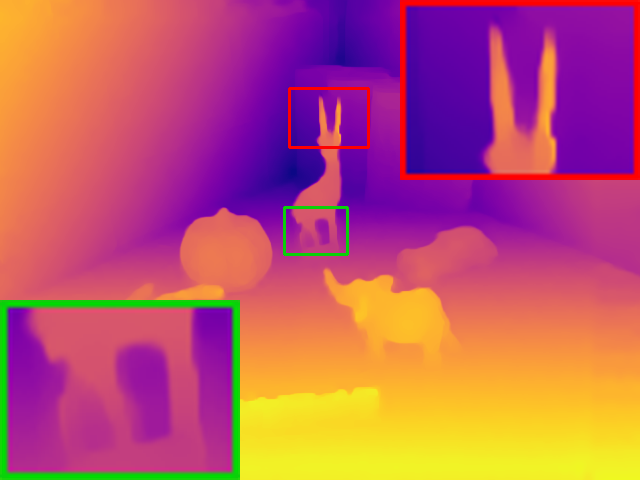}}
    \hfill
    \subcaptionbox{D2A2 (Ours)}{
		\includegraphics[width=0.48\linewidth]{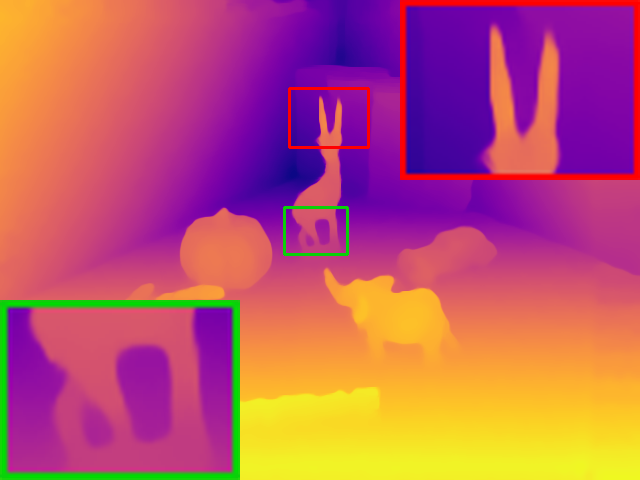}}
\end{center}
\vspace{-6mm}
    \caption{Visual comparison between our D2A2 and the state-of-the-art methods on the Lu dataset~\cite{Lu} for $\times$8 depth super-resolution. We show the enlarged details in the red and green boxes. In contrast, our method can achieve sharper and clearer boundaries than the compared methods. 
    }
\label{fig:effect}
\end{figure}


To employ the  RGB images, existing methods~\cite{hui2016depth, guo2018hierarchical, DJF, PAC, DKN+FDKN, FDSR+RGBDD, DCTNet, DADA} typically incorporate additional branches to extract RGB features and then insert them into depth super-resolution networks.
Such approaches have suggested promise in the recovery of lost details in depth images.
However, they still suffer from issues that lead to noise and blurred boundaries in the super-resolution depth maps. 
The potential reasons that cause these issues can be summarized as follows.
1) The modal misalignment between the depth image and RGB image is neglected in existing methods. This issue arises because depth sensors and RGB cameras operate on different imaging principles.
2) Depth features and RGB features suffer from geometrical misalignment caused by the sensing ability of different sensors. Such a misalignment restrains the accurate cross-modal feature fusion when directly combining both features.
3) Although RGB images and their corresponding depth maps exhibit structural similarities, RGB images also contain significant amount of texture information that is unrelated to the depth map. 
Merely concatenating the extracted RGB features with the depth backbone network for feature filtering and utilization is  suboptimal, which can lead to missing details and texture over-transfer.


Aiming at the aforementioned issues, we propose a Dynamic Dual Alignment and Aggregation network (D2A2) for GDSR. 
To address the modal and geometrical misalignment issues, we propose a dynamic dual alignment module that adapts to alleviate the domain misalignment via a learnable domain alignment block and align cross-modal features geometrically via a dynamic geometrical alignment block by estimating the offset in the deformable convolution~\cite{DCN, DCNv2}. 
The learnable domain alignment block dynamically adjusts the distribution in the feature domain and provides flexibility and powerful representation ability. 
The offset estimation is coupled with the learnable domain alignment block to improve the accuracy of the geometrical alignment. 
The aligned features are then more suitable for subsequent feature aggregation operations.
To address the issue of insufficient filtering and utilization of RGB features,  
we propose a mask-to-pixel feature aggregate module that employs gated convolution~\cite{GateConv} and pixel attention~\cite{PA} to generate effective information from RGB and depth as masked features, and then combine the masked features with depth features.
%
The mask-to-pixel fashion extracts and utilizes masked features while filtering out irrelevant texture noise from RGB features to reduce missing details, 
and avoids introducing artificial artifacts into the results and texture over-transfer.
Our method significantly improves the performance of GDSR over state-of-the-art methods both qualitatively and quantitatively. A set of visual comparisons is shown in Figure \ref{fig:effect}. Our method shows obvious advantages over previous methods, especially in terms of detail and boundary recovery.
%


\textbf{The main contributions are summarized as follows.}
\begin{itemize}
    \setlength{\itemsep}{2pt}
    \setlength{\parsep}{-1pt}
    \item To solve the obstacles of modal and geometrical misalignment in the cross-modal RGB and depth features, we propose a dynamic dual alignment module which dynamically aligns features on both domain and geometry.
    \item We present a mask-to-pixel feature aggregation module to effectively filter and utilize masked features with rich details and clear boundaries, and avoid interference from invalid features. 
    \item We show that simple redesign and integration of cross-modal alignment and aggregation would lead to state-of-the-art GDSR performance and exhibit strong generalization ability in different scenarios and lighting conditions. 
\end{itemize}

\section{Related Work}
\subsection{Guided Depth Super-Resolution}



\begin{figure*}[!t]
\begin{center}
    \includegraphics[width=1\linewidth]{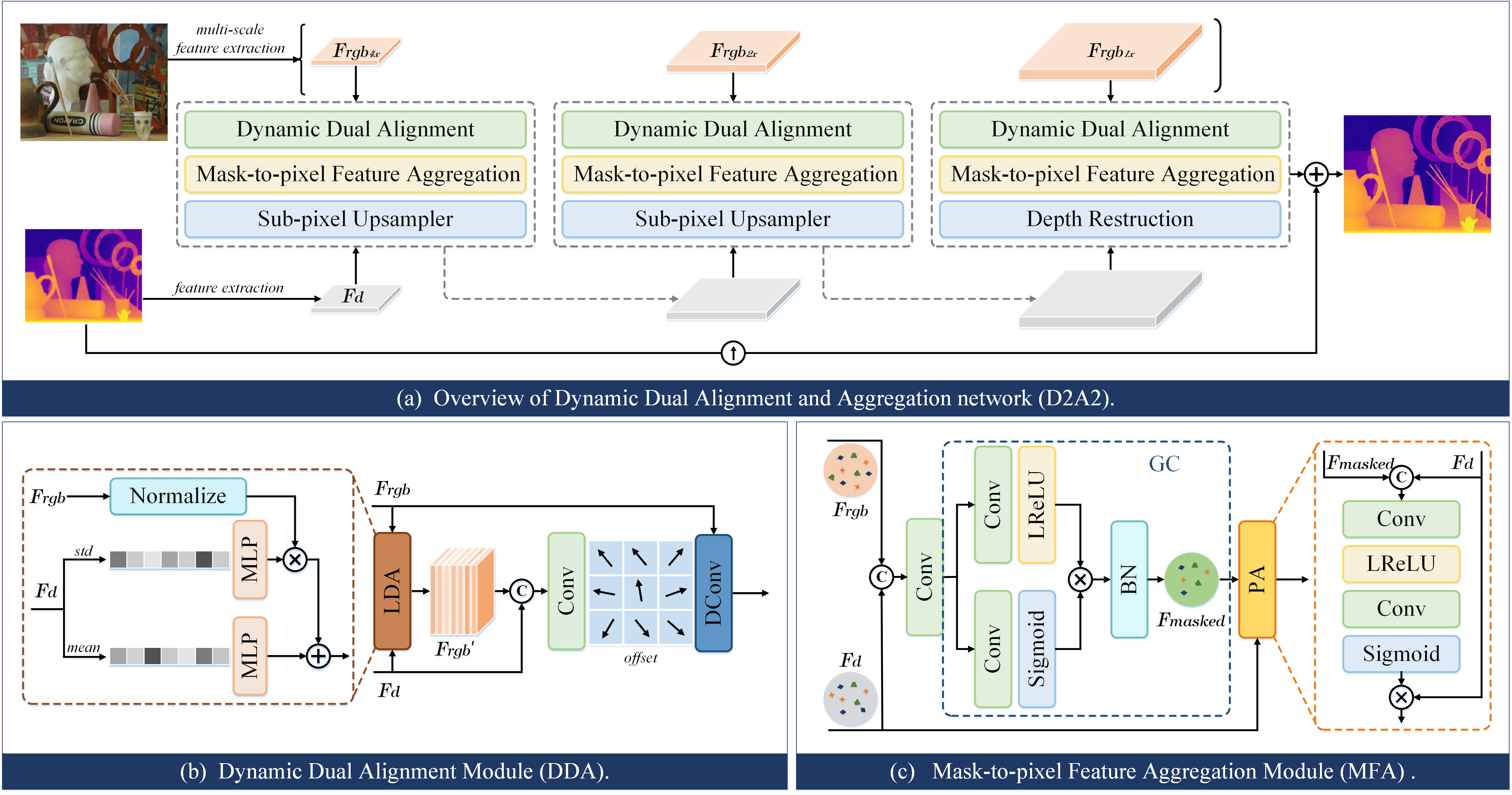}
\end{center}
\vspace{-6mm}
    \caption{(a) is an overview of the proposed D2A2 network. The backbone of D2A2 is a multi-scale architecture with global skip-connection. At each scale, $F_d$ and the corresponding $F_{rgb}$ pass through DDA for modal and geometric alignment, then the aligned $F_{rgb}$ and $F_d$ pass through MFA to obtain the effective feature $F_{masked}$ and fuse it with $F_d$. (b) and (c) are the specific structures of DDA and MFA, respectively.}
\label{fig:model}
\end{figure*}

Machine learning and deep learning bring more possibilities to GDSR \cite{ZhouICCV23,YanNips22,ZhouACMM22,Zhou22ACM,guo2018hierarchical}. 
Li et al.~\cite{DJF, DJFR} proposed a joint filtering method that enhances images degraded due to noise or low spatial resolution by utilizing the similar structure information of the reference image, while avoiding the transmission of some wrong texture information in the reference image that does not originally exist in the reconstructed image, namely texture copy artifacts.
Hui et al.~\cite{hui2016depth} first proposed a multi-scale fusion network framework, which integrates rich color image features from different levels to solve the blurring problem in depth image super-resolution reconstruction.
Guo et al.~\cite{guo2018hierarchical} proposed a layered feature-driven method, which constructs the input pyramid and guidance pyramid of multi-level residual structure to guide the accurate interpolation of low-resolution depth images.
Su et al.~\cite{PAC} proposed a pixel-adaptive convolution method, in which the convolution operation can learn a spatially varying kernel and fuse guidance image features into low-resolution depth image input.
Kim et al.~\cite{DKN+FDKN} proposed a deformable kernel network which adaptively learns a set of sparsely selected neighborhood and interpolation weights through convolutional neural networks and then applies explicit image filters to calculate the final reconstruction result.
He et al.~\cite{FDSR+RGBDD} proposed a fast depth map super-resolution network based on the idea that high-frequency components contain texture details. It adaptively decomposes the high-frequency component from the color reference image to help the super-resolution reconstruction of the depth input image.
Zhao et al.~\cite{DCTNet} introduce a semi-coupled feature extraction module that uses shared convolutional kernels to extract common information and private kernels to extract modality-specific information and employ an edge attention mechanism to highlight the contours informative for guided upsampling. 
Metzger et al.~\cite{DADA} propose a hybrid method based on both optimization and deep learning, which combines guided anisotropic diffusion with a deep convolutional network. 

While the aforementioned methods have yielded favorable results, they overlooked the alignment and effective fusion of RGB image features and depth image features during their integration. We found that the devil is in such details and thus proposed to boost GDSR via rethinking cross-modal alignment and aggregation.

\subsection{Multi-Source Alignment and Fusion}

In reference-based super-resolution (RefSR), alignment and fusion have been incorporated, distinguishing it from most other GDSR methods. 
Wang et al.~\cite{wang2021dual} use an aligned attention module, which searches for related reference patches and warps them to align with the LR counterparts. They also present an adaptive fusion module that aggregates neighbor confidences with an additional convolution net. 
Xia et al.~\cite{xia2022coarse} proposed a multi-scale dynamic aggregation module, which addresses small-scale disalignment through dynamic feature aggregation and reduces large-scale disalignment by fusing multi-scale information.
Cao et al.~\cite{cao2022reference} designed a novel reference-based deformable attention module for correspondence matching and texture transfer. It utilizes a correspondence matching function to calculate the relevance between the Ref and LR images, and a texture transfer function to transfer textures.
Zhang et al.~\cite{zhang2023lmr} use a multi-reference attention module for feature fusion of an arbitrary number of
reference images, and a spatial aware filtering module
for the fused feature selection.
Furthermore, alignment and fusion have been applied successfully in other multimodal tasks, such as medical image registration~\cite{miao2016real, gao2020generalizing, frysch2021novel, liu20222d, markova2022global}, remote sensing image fusion~\cite{zhang2020deeply, chen2021spatiotemporal, li2021saliency,jha2023gaf}, infrared and visible image fusion~\cite{raza2021ir, tang2022image, tang2022piafusion, tang2023divfusion}, and hyperspectral and multispectral image fusion~\cite{zhang2020ssr, zheng2021edge, sun2022mlr}.
Different from the above methods, we propose the idea of aligning from both modal and geometric aspects, and using the gating mechanism to obtain effective features before fusion. 

\section{Methodology}

\begin{figure*}[!t]
\begin{center}
    \subcaptionbox{RGB}{
		\includegraphics[width=0.16\linewidth]{images/paper/mask_nyu/RGB.jpg}}
    \hspace{-5px}
    \subcaptionbox{Ground Truth}{
		\includegraphics[width=0.16\linewidth]{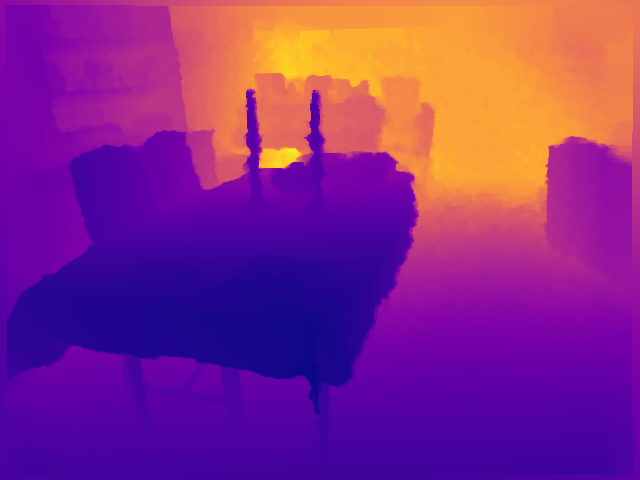}}
    \hspace{-5px}
    \subcaptionbox{Before DDA}{
		\includegraphics[width=0.16\linewidth]{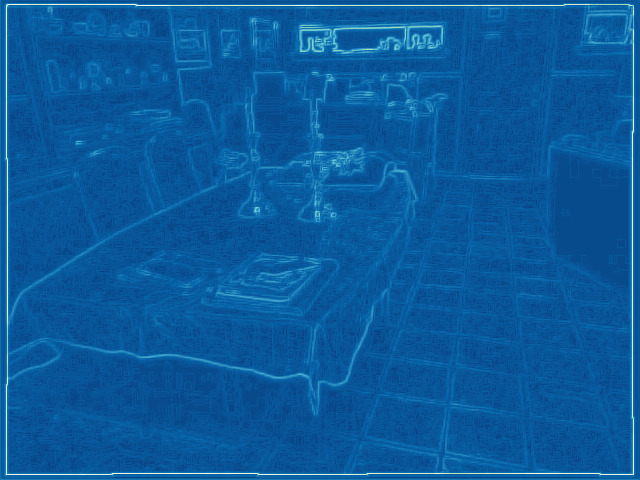}}
    \hspace{-5px}
    \subcaptionbox{After DDA}{
		\includegraphics[width=0.16\linewidth]{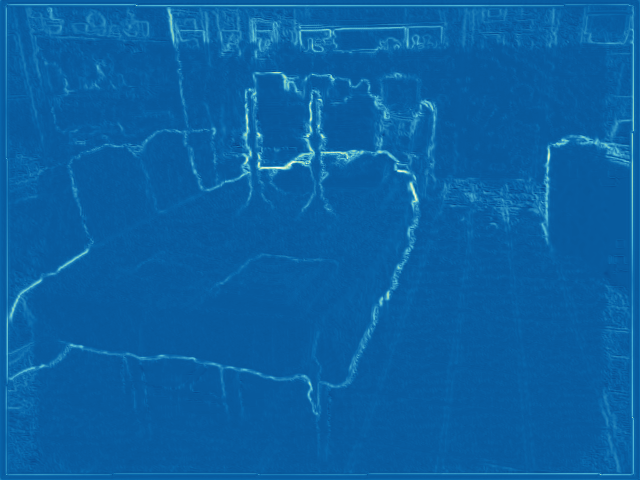}}
    \hspace{-5px}
    \subcaptionbox{Mask of GC}{
		\includegraphics[width=0.16\linewidth]{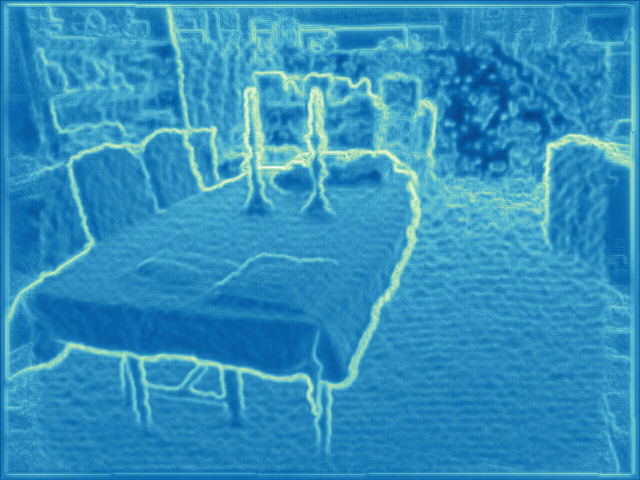}}
    \hspace{-5px}
    \subcaptionbox{Weight map of PA}{
		\includegraphics[width=0.16\linewidth]{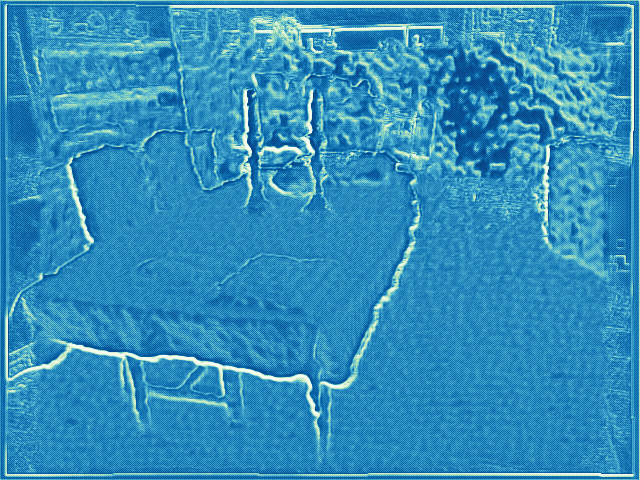}}
\end{center}
\vspace{-6mm}
    \caption{Visualization of RGB features before and after the dynamic dual alignment module (DDA) as well as the mask of the gated convolution (GC) and the weight map of pixel attention (PA) in the mask-to-pixel aggregation module.}
\label{fig:aligned_mask}
\end{figure*}

Figure \ref{fig:model}(a) presents an overview of our D2A2 framework.
The depth super-resolution network primarily consists of two core modules: the dynamic dual alignment (DDA) module and the mask-to-pixel feature aggregation (MFA) module. We detail the core modules of our method as follows.


\subsection{Dynamic Dual Alignment Module}

The dynamic dual alignment module is proposed to solve the modal and geometrical misalignment issues in the cross-modal RGB and depth features. As presented in Figure \ref{fig:model}(b), it comprises a learnable domain alignment block (LDA) and a dynamic geometrical alignment block (DGA). 
Specifically, the DDA module takes the RGB features and depth features as input.
We first utilize the LDA to align the cross-modal features, which adjusts the distribution of two different modalities in the feature domain.
Then, we concatenate the depth and aligned RGB features along the channel dimension. 
We employ a convolution layer to predict the offset of the RGB features relative to the depth features.
Finally, we use DGA to adjust the RGB features based on the predicted offset.

Figure \ref{fig:aligned_mask}(c) and (d) visualize the RGB feature before and after the dynamic dual alignment module.
It is evident that the lines of the tablecloth and candlestick after DDA are more in line with the ground truth (Figure \ref{fig:aligned_mask}(b)) than RGB (Figure \ref{fig:aligned_mask}(a)). The comparison reveals that the crisp and sleek straight lines and rectangles in RGB transform into irregular and distorted shapes after the DDA module, i.e., the RGB feature after the DDA module is more compatible with the depth ground truth.

\begin{table*}[!t]
\caption{Quantitative comparisons in terms of the RMSE under different upscaling factors. The best and second
performances are marked in \textbf{bold} and \underline{underlined}, respectively. 
}
\vspace{-6mm}
	\begin{center}
		\begin{tabular}{ccccccccccccc}
			\hline
			\multirow{2}{*}{Methods} & \multicolumn{3}{c}{Middlebury} & \multicolumn{3}{c}{Lu} & \multicolumn{3}{c}{NYUv2} & \multicolumn{3}{c}{RGBDD}  \\ 
			\cmidrule(lr){2-4} \cmidrule(lr){5-7} \cmidrule(lr){8-10} \cmidrule(lr){11-13}
			\makebox[0.06\textwidth]{}& \makebox[0.045\textwidth]{×4} & \makebox[0.045\textwidth]{×8} & \makebox[0.045\textwidth]{×16} & \makebox[0.045\textwidth]{×4} & \makebox[0.045\textwidth]{×8} & \makebox[0.045\textwidth]{×16} & \makebox[0.045\textwidth]{×4} & \makebox[0.045\textwidth]{×8} & \makebox[0.045\textwidth]{×16} & \makebox[0.045\textwidth]{×4} & \makebox[0.045\textwidth]{×8} & \makebox[0.045\textwidth]{×16} \\
			\hline\hline
			DJF~\cite{DJF} & 1.68& 3.24& 5.62 &
			1.65& 3.96& 6.75 &
			2.80& 5.33& 9.46 &
			3.41& 5.57& 8.15 \\
			DJFR~\cite{DJFR} & 1.32& 3.19& 5.57 &
			1.15& 3.57& 6.77& 
			2.38& 4.94& 9.18& 
			3.35& 5.57& 7.99 \\
			PAC~\cite{PAC} & 1.32& 2.62& 4.58& 
			1.20& 2.33& 5.19&
			1.89& 3.33& 6.78&
			1.25& 1.98& 3.49  \\
			CUNet~\cite{CUNet} & 1.10& 2.17& 4.33&
			0.91& 2.23& 4.99& 
			1.92& 3.70& 6.78&
			1.18& 1.95& 3.45 \\
			DKN~\cite{DKN+FDKN} & 1.23& 2.12& 4.24& 
			0.96& 2.16& 5.11&
			1.62& 3.26& 6.51&
			1.30& 1.96& 3.42 \\
			FDKN~\cite{DKN+FDKN} & \underline{1.08}& 2.17& 4.50&
			\underline{0.82}& 2.10& 5.05&
			1.86& 3.58& 6.96&
			1.18& 1.91& 3.41 \\
			FDSR~\cite{FDSR+RGBDD} & 1.13& 2.08& 4.39&
			1.29& 2.19& 5.00&
			1.61& 3.18& 5.86& 
			1.16& 1.82& 3.06 \\
			DCTNet~\cite{DCTNet} & 1.10& 2.05& 4.19&
			0.88& \underline{1.85}& 4.39&
			1.59& 3.16& 5.84&
			\textbf{1.08}& \underline{1.74}& 3.05 \\
            DADA~\cite{DADA} & 1.20& \underline{2.03}& \underline{4.18}&
			0.96& 1.87& \underline{4.01}&
			1.54& 2.74& \textbf{4.80}&
			1.20& 1.83& \underline{2.80} \\
			D2A2 (Ours) & \textbf{1.04}&\textbf{1.67}&\textbf{3.26}&
			\textbf{0.81}&\textbf{1.54}&\textbf{3.80}&
			\textbf{1.30}&\textbf{2.62}&\underline{5.12}&
			\underline{1.11}&\textbf{1.72}&\textbf{2.72}  \\
			\hline
		\end{tabular}
	\end{center}
    \label{tab:rmse}
\end{table*}

\begin{figure}[!t]
\begin{center}
    \includegraphics[width=1\linewidth]{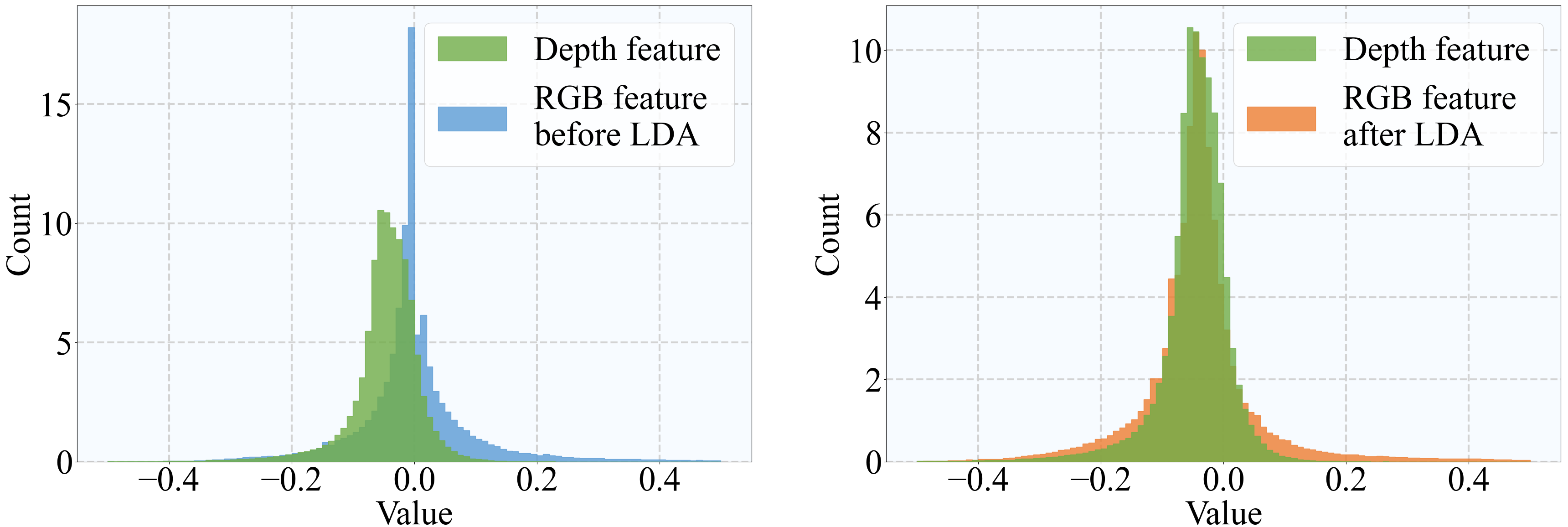}
\end{center}
\vspace{-6mm}
    \caption{The histogram of the input RGB feature and RGB feature after LDA with corresponding the histogram of depth feature. The distribution of the RGB feature after LDA is more inclined towards the depth feature. }
\label{fig:c}
\end{figure}

\noindent
\textbf{Learnable domain alignment.} 
To mitigate the impact of modal misalignment between RGB features $F_{rgb}$ and depth features $F_{d}$, we design a learnable domain alignment block (LDA) with learnable parameters to adjust the data distribution of RGB features to depth features, which can control the degree of migration adaptively. LDA separately forwards the mean and standard deviation of the depth features to the MLP modules to estimate the optimal parameters. After normalizing RGB features, the mean and standard deviation of $F_{d}$ after MLP are used for denormalization, and the obtained result are RGB features aligned towards depth. 
It can be expressed as
\begin{gather}
\sigma_{F_{d}}' = \text{MLP}(\sigma_{F_{d}}), \nonumber\\
\mu_{F_{d}}' = \text{MLP}(\mu_{F_{d}}), \nonumber\\
LDA(F_{rgb}, F_{d}) = 
    \sigma_{F_{d}}' \left( \frac{F_{rgb} - \mu_{F_{rgb}}}{\sigma_{F_{rgb}}}   \right)
    + \mu_{F_{d}}',
\end{gather}
where $\mu$ and $\sigma$ are the mean and standard divination, respectively.
LDA can dynamically adjust the distribution of feature domains with strong flexibility and representation ability rather than rigid normalization, as shown in Figure \ref{fig:c}. 

\noindent
\textbf{Dynamic geometrical alignment.}
Motivated by the ability of deformable convolution~\cite{DCNv2} in geometric transformations, we use it to align the features of RGB and depth geometrically. 
Specifically,  the dynamic weight map of the sampling core is learned with a modulated scalar. Modulated deformable convolution learns the modulation scalar and the offset jointly, enabling the kernel to have greater spatial variability. Mathematically, the deformable convolution operation can be expressed as
\begin{equation}
    Y(p) = \sum_{k=1}^{K} w_k \cdot X(p+p_k+\Delta p_k) \cdot \Delta m_k,
\end{equation}
where $X$ denotes input features, $Y$ denotes output features, and $k$ and $K$ represent the number of index and kernel weights, respectively. $w_k$, $p$, $p_k$, and $\Delta p_k$ are the weights of the $k$th kernel, the center index, the $k$th fixed offset, and the learnable offset of the $k$th position, respectively, $p_k \in \{(-1,1), (-1,0), ..., (1,1)\} $. $\Delta m_k$ is the modulation scalar.
We estimate the offset using the features after LDA and apply it to wrap RGB features to depth features.

\subsection{Mask-to-Pixel Feature Aggregation Module}

After aligning the RGB features with depth features by the DDA module, we utilize a mask-to-pixel feature aggregation (MFA) module to adaptively filter and fuse the cross-modal features. 
We introduce the gating mechanism and pixel attention mechanism in the MFA module. The input RGB features and depth features are represented as $F_{rgb}\in R^{h\times w \times c}$ and $F_{d}\in R^{h\times w \times c}$. Through gated convolution (GC), we get masked features $F_{masked}\in R^{h\times w \times c}$, then $F_{masked}$ is used to guide the reconstruction of $F_d$ through pixel attention (PA). The detailed structures are illustrated in Figure \ref{fig:model}(c).


\noindent
\textbf{Gated Convolution.}
One of the key differences between RGB and depth is that RGB contains more details such as edges and textures than depth, which may not be relevant to depth information. 
For instance, when a box is placed in a scene, the patterns on the box are irrelevant to the depth, but the box border is the crucial cue for depth estimation. 
To discard the irrelevant information when fusing cross-modal features, we introduce gated convolution~\cite{GateConv} that can adaptively update the mask to separate effective pixels from invalid pixels.
As illustrated in Figure \ref{fig:model}(c), gated convolution consists of two parallel convolutional layers: a ``feature'' convolution and a ``gate'' convolution. 
The feature convolution performs a standard convolution operation on the input, while the gate convolution generates a gating signal that controls which parts of the input are transmitted to the output. 
Specifically, the gate convolution produces a set of activation values between 0 and 1, which are multiplied element-wise with the output of the feature convolution.

The confidence diagram of the mask in the gated convolution is presented in Figure \ref{fig:aligned_mask}(e). 
The lighter the color, the higher its confidence. Figure \ref{fig:aligned_mask}(e) demonstrates that the edges related to depth, such as the outlines of chairs and the table, are clearly distinguishable due to their light coloration, indicating that GC has effectively identified them as valid features. Conversely, parts of the image that are irrelevant to depth, such as the lines of floor tiles, appear relatively dark and less distinguishable, indicating that GC has identified them as invalid features.
The visualization result suggests the effectiveness of gated convolution in distinguishing useful and irrelevant features.

\begin{figure*}
\begin{center}
    \subcaptionbox{RGB}{
		\includegraphics[width=0.19\linewidth]{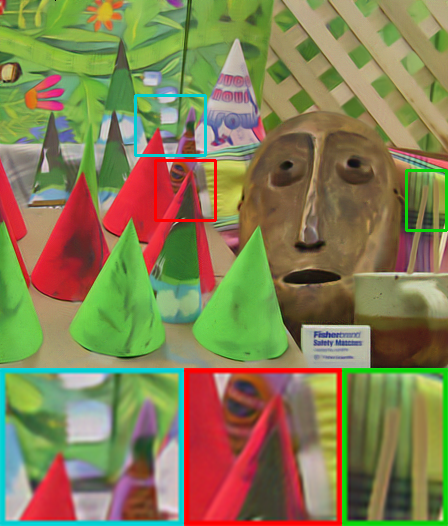}}
    \hfill
    \subcaptionbox{Ground Truth}{
		\includegraphics[width=0.19\linewidth]{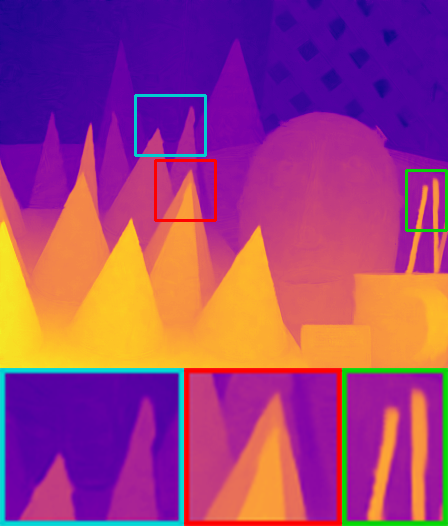}}
    \hfill
    \subcaptionbox{PAC~\cite{PAC}}{
        \includegraphics[width=0.19\linewidth]{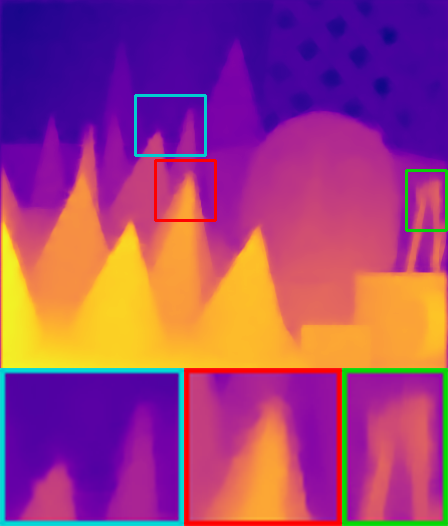}}
    \hfill
    \subcaptionbox{DKN~\cite{DKN+FDKN}}{
        \includegraphics[width=0.19\linewidth]{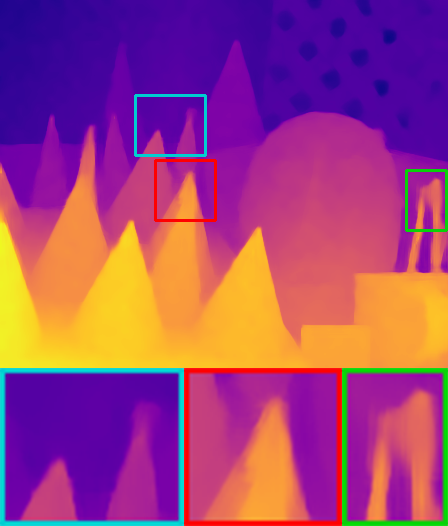}}
    \hfill
    \subcaptionbox{FDKN~\cite{DKN+FDKN}}{
        \includegraphics[width=0.19\linewidth]{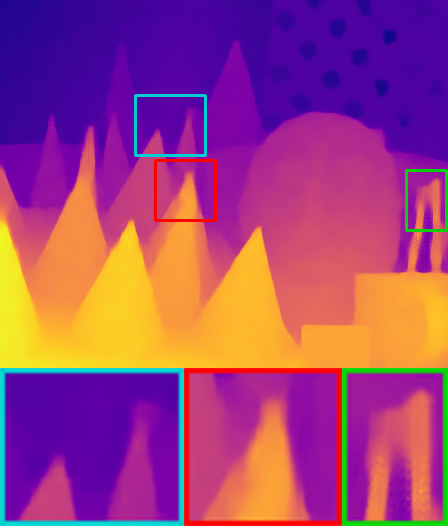}}
    \\
    \subcaptionbox{CUNet~\cite{CUNet}}{
		\includegraphics[width=0.19\linewidth]{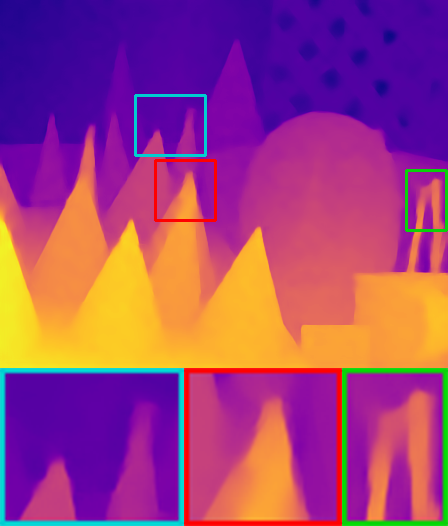}}
    \hfill
    \subcaptionbox{FDSR~\cite{FDSR+RGBDD}}{
		\includegraphics[width=0.19\linewidth]{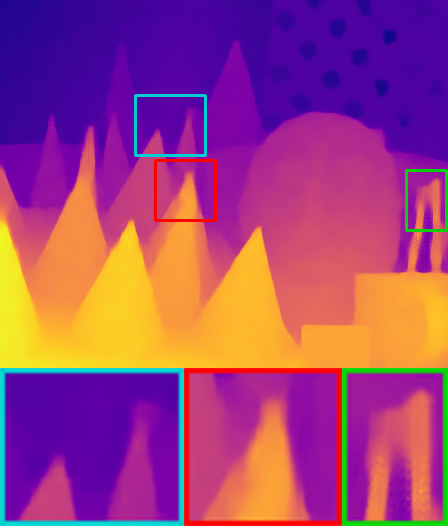}}
    \hfill
    \subcaptionbox{DCTNet~\cite{DCTNet}}{
        \includegraphics[width=0.19\linewidth]{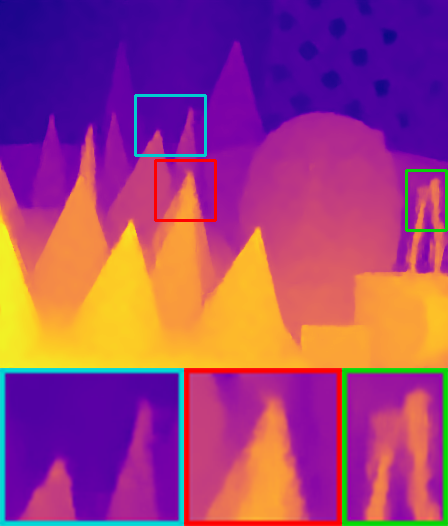}}
    \hfill
    \subcaptionbox{DADA~\cite{DADA}}{
        \includegraphics[width=0.19\linewidth]{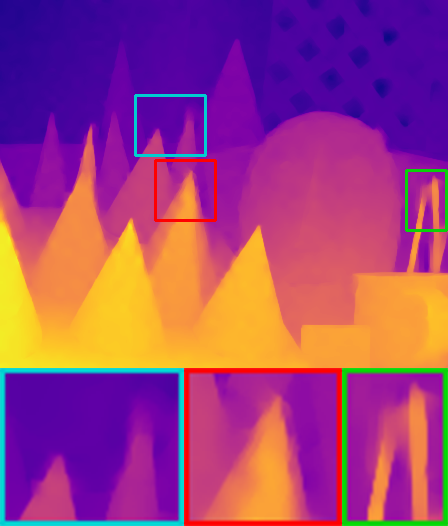}}
    \hfill
    \subcaptionbox{D2A2 (Ours)}{
        \includegraphics[width=0.19\linewidth]{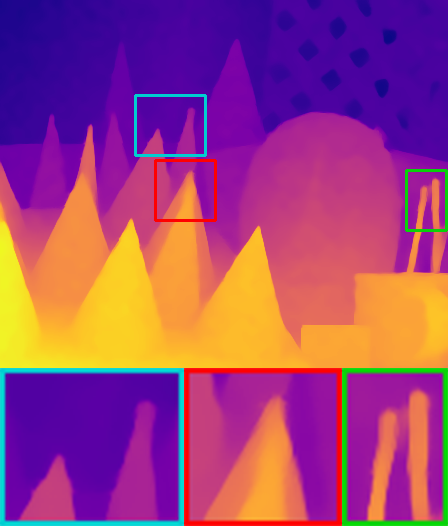}}
\end{center}
\vspace{-6mm}
\caption{Visual comparison of different methods on the    Middlebury dataset for $\times$8 depth super-resolution. }
\label{fig:sr_m}
\end{figure*}


\noindent
\textbf{Pixel Attention.}
Previous methods typically use simple fusion techniques to fuse the RGB and depth features without measuring the feature quality. 
These arbitrary fusion operations may ``copy'' irrelevant or noisy information from the RGB image.
To solve this issue, we employ pixel attention~\cite{PA}  to fuse the depth features and the output features of the gated convolution.
Unlike the spatial and channel attention mechanisms, the pixel attention mechanism generates a three-dimensional matrix as an attention feature. 
This mechanism is intrinsically suitable for the case where the depth and RGB images have strong structural similarities. 
This is because pixel attention can pay attention to important details pixel by pixel during the feature fusion process.

The pixel attention mechanism shown in Figure \ref{fig:model}(c) operates as follows. 
Firstly, the input features and guide features are connected along the channel dimension, and then the dimension is reduced through convolution to match the input feature's dimensions. 
Next, the Sigmoid function is applied to obtain the attention map. 
Finally, the attention map is multiplied by the input features to produce the pixel attention enhancement output. 
The confidence diagram of the weight map of pixel attention is shown in Figure \ref{fig:aligned_mask}(f). 
Observing Figure \ref{fig:aligned_mask}(f), we can see the presence of rich details, and the varying weights assigned to different spatial coordinates. This suggests that PA can selectively guide the fusion of features between RGB features and depth features to different extents.

\begin{figure*}
\begin{center}
    \subcaptionbox{RGB}{
		\includegraphics[width=0.19\linewidth]{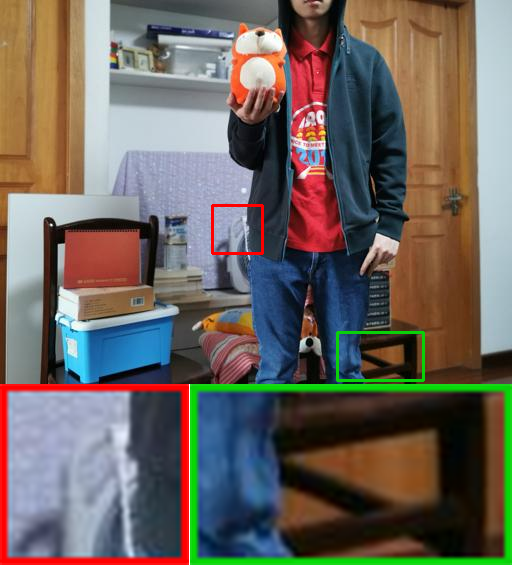}}
    \hfill
    \subcaptionbox{Ground Truth}{
		\includegraphics[width=0.19\linewidth]{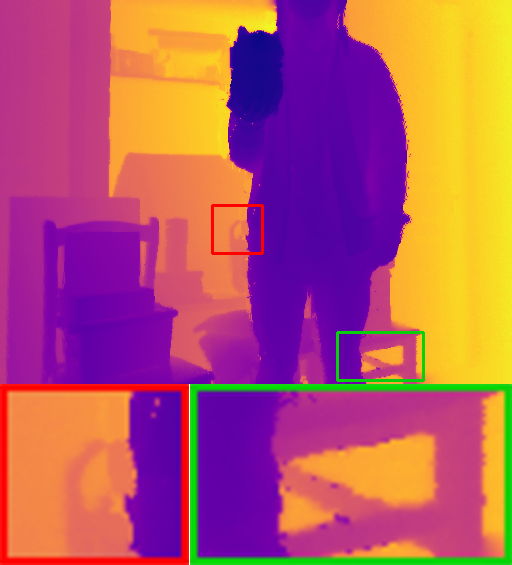}}
    \hfill
    \subcaptionbox{PAC~\cite{PAC}}{
        \includegraphics[width=0.19\linewidth]{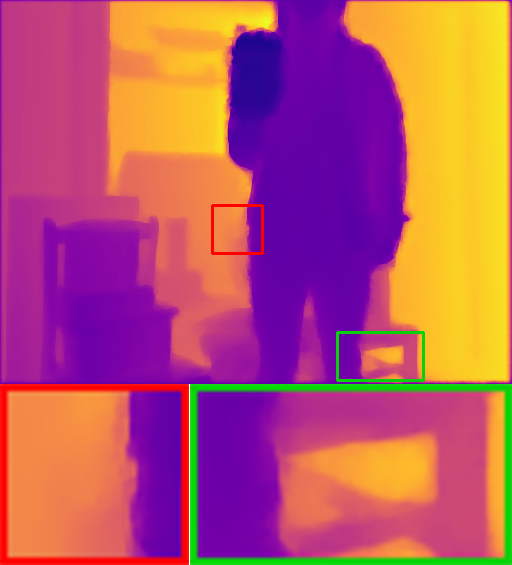}}
    \hfill
    \subcaptionbox{DKN~\cite{DKN+FDKN}}{
        \includegraphics[width=0.19\linewidth]{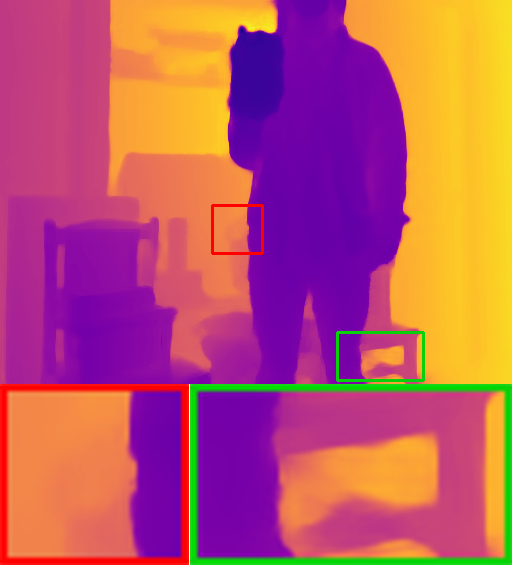}}
    \hfill
    \subcaptionbox{FDKN~\cite{DKN+FDKN}}{
        \includegraphics[width=0.19\linewidth]{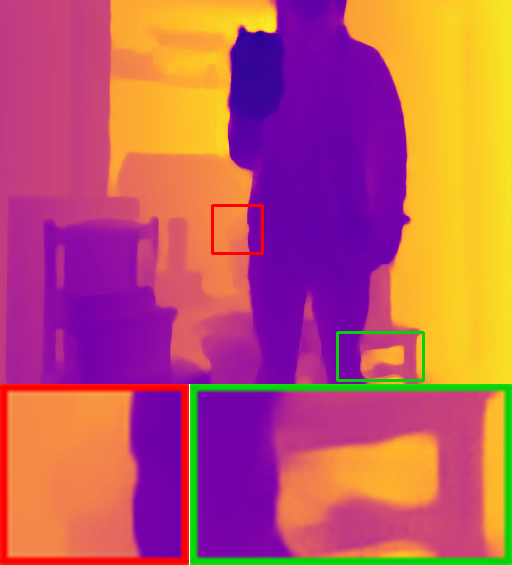}}
    \\
    \subcaptionbox{CUNet~\cite{CUNet}}{
		\includegraphics[width=0.19\linewidth]{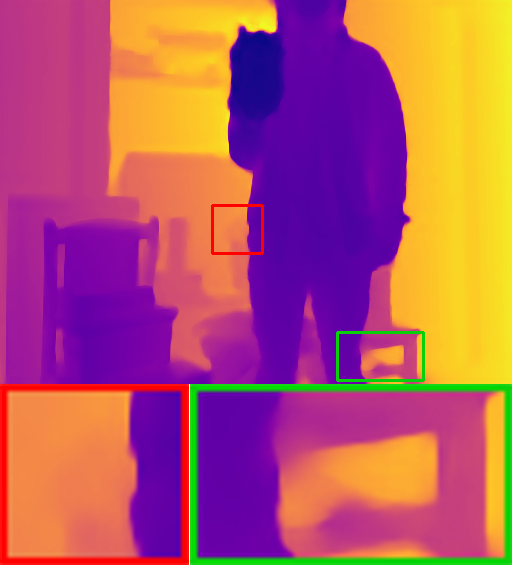}}
    \hfill
    \subcaptionbox{FDSR~\cite{FDSR+RGBDD}}{
		\includegraphics[width=0.19\linewidth]{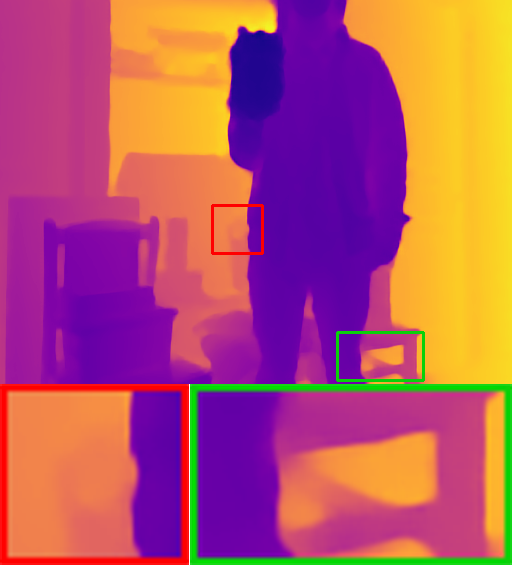}}
    \hfill
    \subcaptionbox{DCTNet~\cite{DCTNet}}{
        \includegraphics[width=0.19\linewidth]{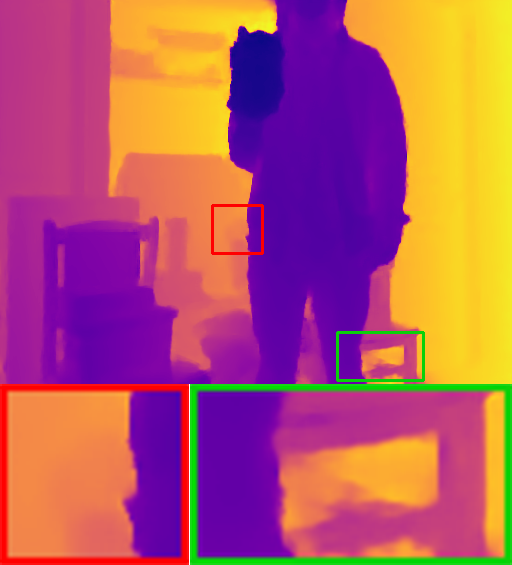}}
    \hfill
    \subcaptionbox{DADA~\cite{DADA}}{
        \includegraphics[width=0.19\linewidth]{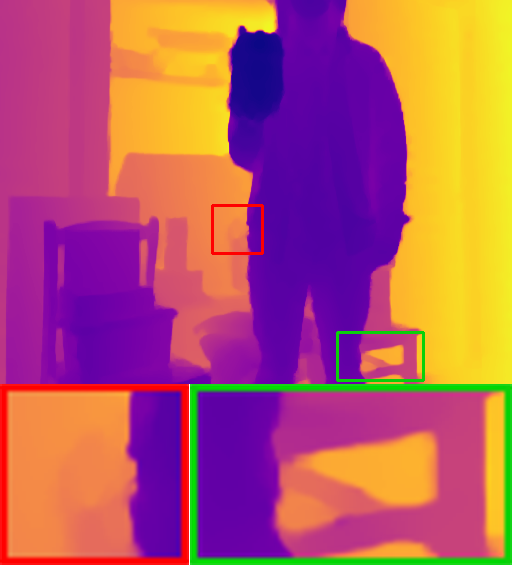}}
    \hfill
    \subcaptionbox{D2A2 (Ours)}{
        \includegraphics[width=0.19\linewidth]{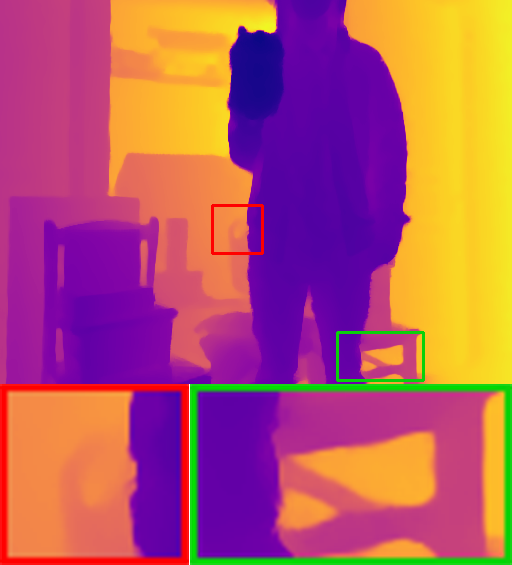}}
\end{center}
\vspace{-6mm}
\caption{Visual comparison of different methods on the RGBDD dataset for $\times$8 depth super-resolution.}
\label{fig:sr_rgbdd}
\end{figure*}

\begin{figure*}
\begin{center}
    \subcaptionbox{RGB}{
		\includegraphics[width=0.19\linewidth]{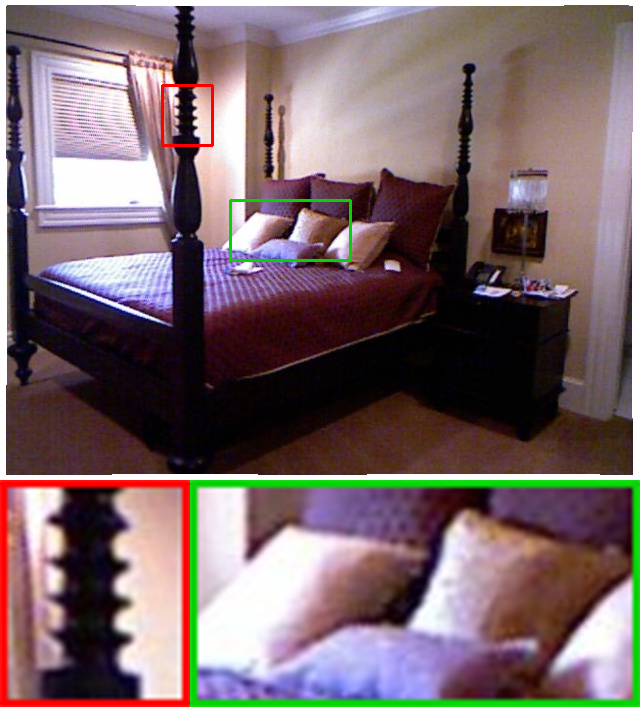}}
    \hfill
    \subcaptionbox{Ground Truth}{
		\includegraphics[width=0.19\linewidth]{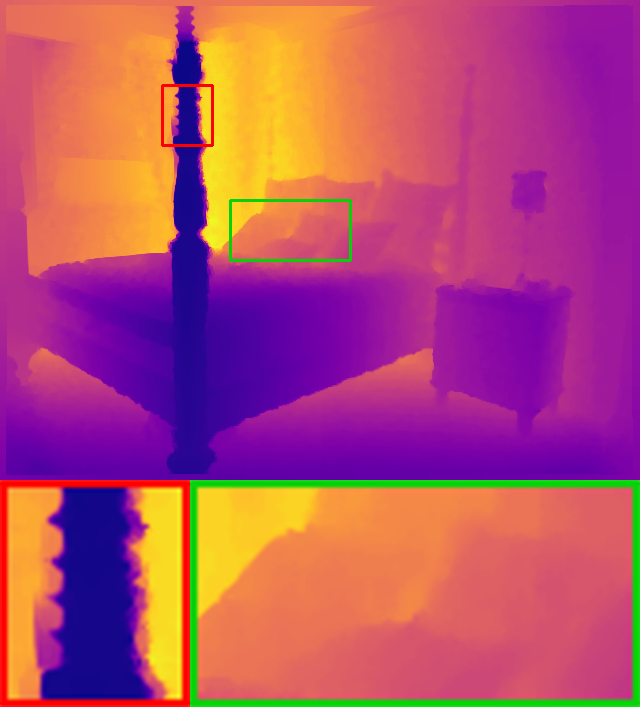}}
    \hfill
    \subcaptionbox{PAC~\cite{PAC}}{
        \includegraphics[width=0.19\linewidth]{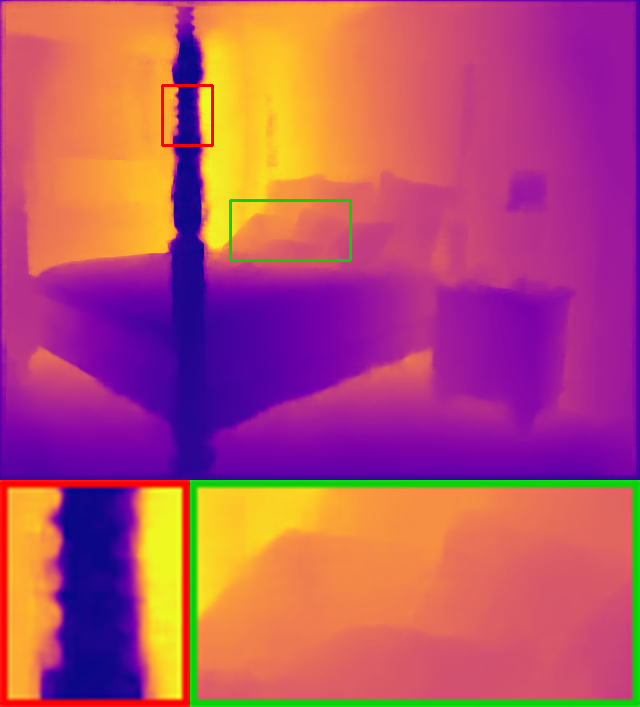}}
    \hfill
    \subcaptionbox{DKN~\cite{DKN+FDKN}}{
        \includegraphics[width=0.19\linewidth]{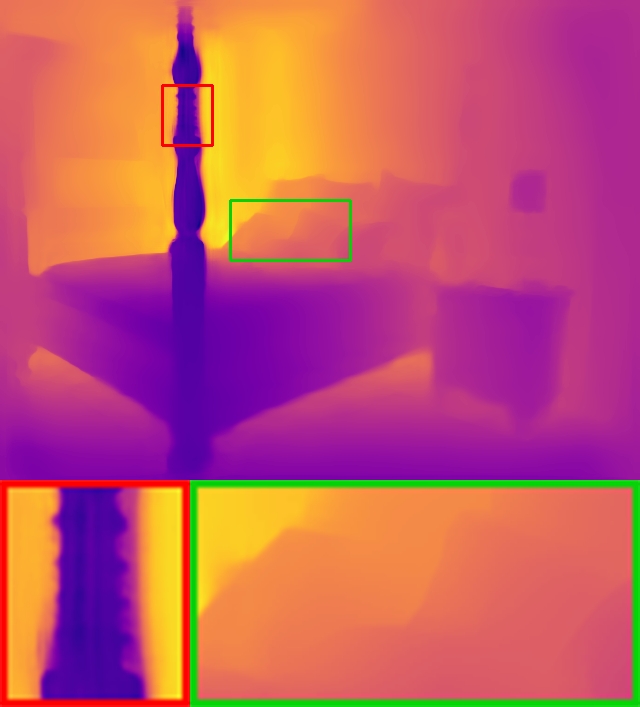}}
    \hfill
    \subcaptionbox{FDKN~\cite{DKN+FDKN}}{
        \includegraphics[width=0.19\linewidth]{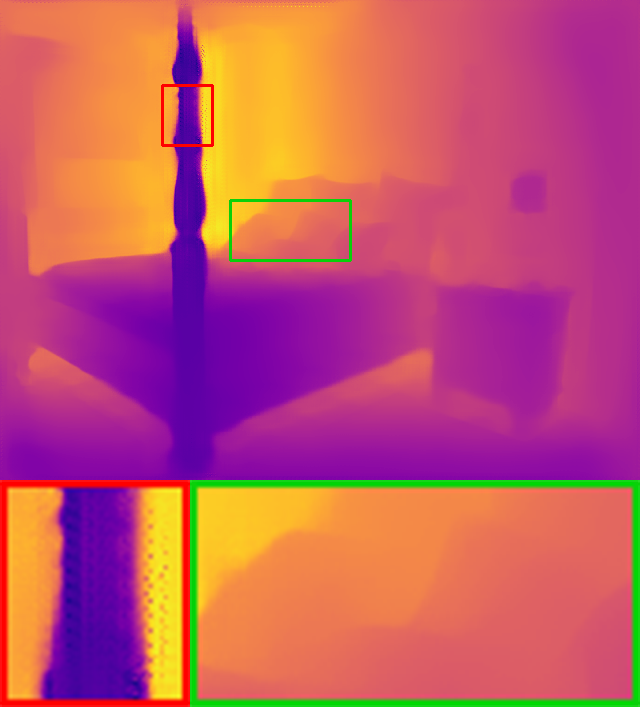}}
    \\
    \subcaptionbox{CUNet~\cite{CUNet}}{
		\includegraphics[width=0.19\linewidth]{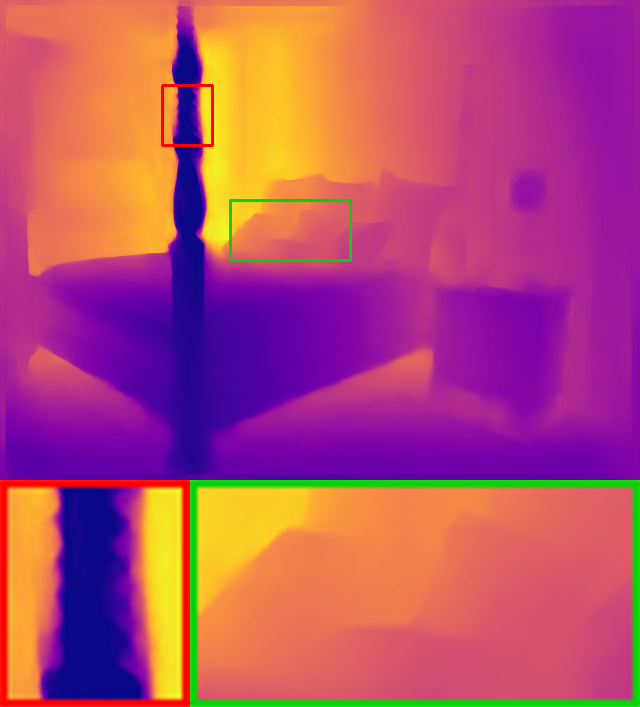}}
    \hfill
    \subcaptionbox{FDSR~\cite{FDSR+RGBDD}}{
		\includegraphics[width=0.19\linewidth]{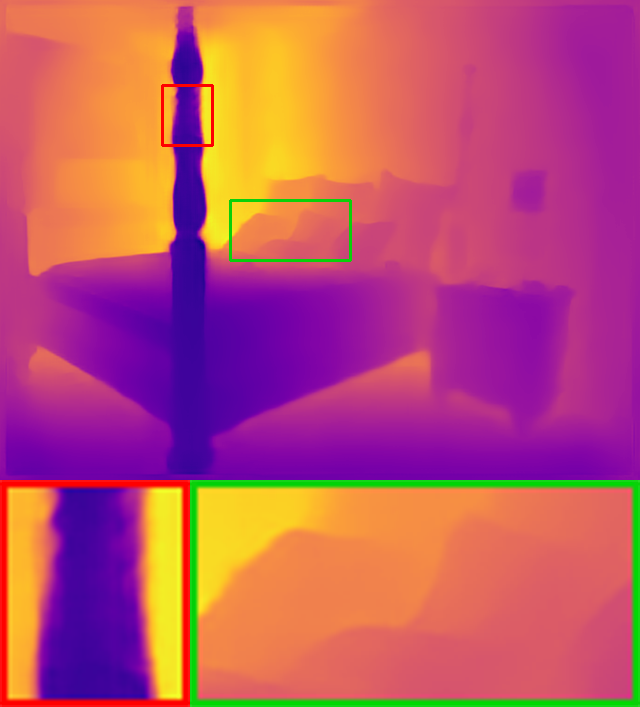}}
    \hfill
    \subcaptionbox{DCTNet~\cite{DCTNet}}{
        \includegraphics[width=0.19\linewidth]{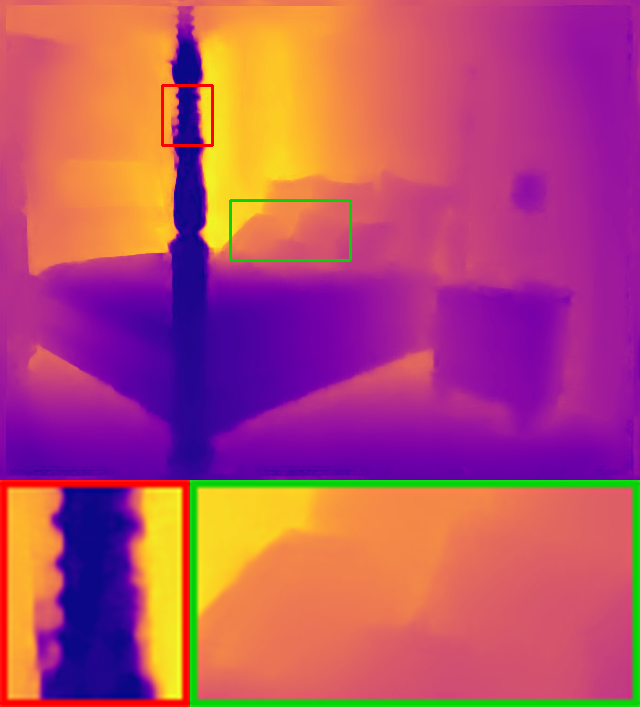}}
    \hfill
    \subcaptionbox{DADA~\cite{DADA}}{
        \includegraphics[width=0.19\linewidth]{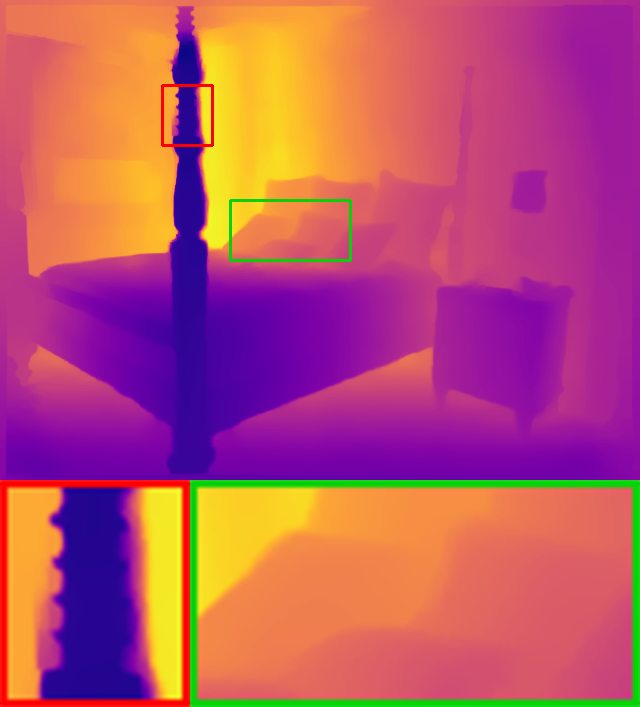}}
    \hfill
    \subcaptionbox{D2A2 (Ours)}{
        \includegraphics[width=0.19\linewidth]{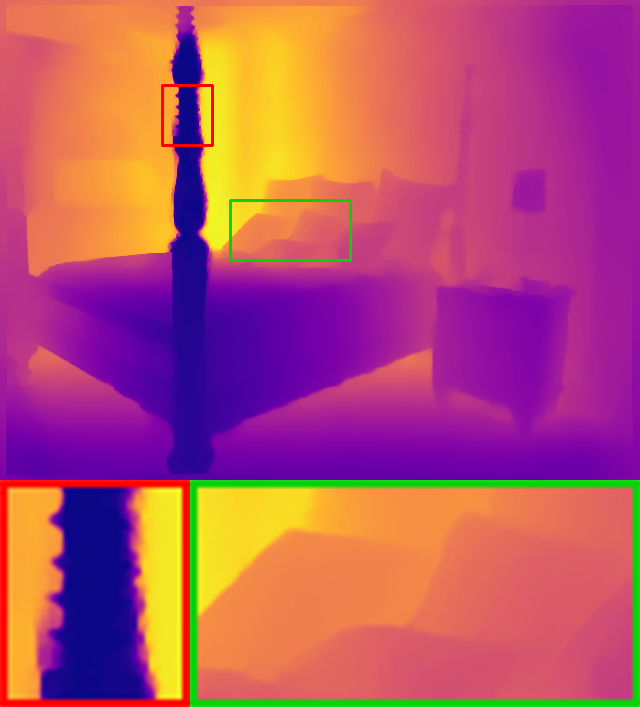}}
\end{center}
\vspace{-6mm}
\caption{Visual comparison of different methods on the NYUv2 dataset for $\times$16 depth super-resolution.}
\label{fig:sr_nyu}
\end{figure*}

\section{Experiment}
\subsection{Experimental Settings}

\noindent
\textbf{Datasets.} 
There are four widely used datasets for GDSR tasks: NYUv2~\cite{NYUv2}, Middlebury~\cite{Middlebury}, Lu~\cite{Lu}, and RGBDD~\cite{FDSR+RGBDD}. 
Specifically, the RGB images of Lu are low-lighting. 
In line with prior research~\cite{DKN+FDKN, FDSR+RGBDD, DCTNet}, we utilize the NYUv2 dataset to train our model. 
Specifically, the first 1,000 pairs of RGB and depth images are used for training, while the remaining 449 pairs are for evaluation.
We also test the generalization performance of our model on the Middlebury, Lu, and RGBDD datasets.
For quantitative evaluation, following previous methods~\cite{DJF, PAC, FDSR+RGBDD, DCTNet}, we choose the Root Mean Square Error (RMSE).

\noindent
\textbf{Training Details.}
We down-sample the HR depth images by bicubic and the training sample pairs are cropped to 256x256 randomly with data augmentation techniques (random horizontal or vertical flips and rotations) and separate data normalization.
The model is trained for 500 epochs with the Adam optimizer~\cite{Adam}, the fixed learning rate of 0.001, and the batch size of 4. 
We employ the L1 Loss as our loss function.
Our implementation is with PyTorch.

\noindent
\textbf{Compared Methods.}
We compare the performance of our method with several state-of-the-art RGB-guided depth image super-resolution methods, 
including DJF (ECCV'16)~\cite{DJF}, 
DJFR (TPAMI'19)~\cite{DJFR}, 
PAC (CVPR'19)~\cite{PAC}, 
CUNet (TPAMI'20)~\cite{CUNet}, 
DKN (IJCV'21)~\cite{DKN+FDKN}, 
FDKN (IJCV'21)~\cite{DKN+FDKN}, 
FDSR (CVPR'21)~\cite{FDSR+RGBDD}, 
DCTNet (CVPR'22)~\cite{DCTNet}, 
and DADA (CVPR'23)~\cite{DADA}.

\subsection{Comparison with the State of the Arts}

\noindent
\textbf{Quantitative Comparison.}
Table \ref{tab:rmse} presents the average RMSE values of different methods on four datasets, namely Middlebury, Lu, NYUv2, and RGBDD, at upscaling factors of $\times$4, $\times$8, and $\times$16. For the NYUv2 dataset, the RMSE values are reported in centimeters, whereas for the Middlebury and Lu datasets, they are reported in the original scale of the provided parallax. For the RGBDD dataset, the RMSE values are measured in meters.
As shown in Table \ref{tab:rmse}, our proposed D2A2 outperforms all the other methods, except for two cases where it ranks second. 
These results demonstrate that the performance of D2A2 is superior to state-of-the-art methods. 
Furthermore, D2A2 exhibits excellent generalization performance in different scenarios and lighting conditions, effectively handling data from other datasets, such as Middlebury, Lu, and RGBDD.

\noindent
\textbf{Qualitative Comparison.}
Figure \ref{fig:sr_m}, \ref{fig:sr_rgbdd} and \ref{fig:sr_nyu} shows the visual comparison of different methods on the Middlebury, RGBDD and NYUv2 dataset for $\times$8, $\times$8 and $\times$16 upscaling, respectively. 
We can observe that our proposed method can produce more accurate details and sharp depth boundaries in the reconstructed depth map, compared to the state-of-the-art methods. In addition, our results show fewer artificial artifacts and noise than the results of other methods, which are commonly caused by the RGB reference image.
This is owing to the effectiveness of our dynamic dual alignment module based on the learnable domain alignment and dynamic geometrical alignment as well as the mask-to-pixel feature aggregation module based on gated convolution and pixel attention.


\subsection{Ablation Studies}
To investigate the impact of core modules in the proposed D2A2 network, we conduct ablation experiments on the NYUv2 dataset at upscaling factors of $\times$4, $\times$8, and $\times$16. 
Baseline refers to D2A2 without DDA and replaces the MFA with a simple concatenation operation.

\begin{table}
\caption{Quantitative results of the ablation studies for the DDA and MFA modules. \textbf{Bold} indicates the best score in terms of RMSE. 
}
\vspace{-6mm}	
 \begin{center}
		\begin{tabular}{c|c|c|c}
			\hline
			Model & $\times$4 & $\times$8 & $\times$16\\
			\hline\hline
			baseline & 1.54 & 3.09 & 6.07 \\
			w/o MFA  & 1.32 & 2.67 & 5.19\\
			w/o DDA & 1.33 & 2.69 & 5.27\\
			D2A2 (Ours) & \textbf{1.30} & \textbf{2.62} & \textbf{5.12} \\
			\hline
		\end{tabular}
	\end{center}
 \vspace{-3mm}
    \label{tab:LADCN_GCPA_moudle}
\end{table}

\begin{figure}
\begin{center}
    \subcaptionbox{RGB}{
		\includegraphics[width=0.23\linewidth]{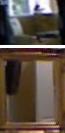}}
    \hfill
    \subcaptionbox{GT}{
		\includegraphics[width=0.23\linewidth]{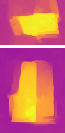}}
    \hfill
    \subcaptionbox{w/o DDA}{
		\includegraphics[width=0.23\linewidth]{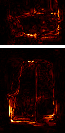}}
    \hfill
    \subcaptionbox{w/ DDA}{
		\includegraphics[width=0.23\linewidth]{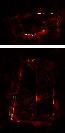}}
\end{center}
\vspace{-6mm}
    \caption{The geometric part of error maps of w/ or w/o DDA on the NYUv2 dataset for $\times$4 depth super-resolution. (d) has lower prediction errors than (c), which indicates that DDA performs well in aligning geometry. The sizes of the cropped regions are 45$\times$65 and 85$\times$65 pixels, respectively. 
    }
\label{fig:errormap_DDA}
\end{figure}

\begin{figure}
\begin{center}
    \subcaptionbox{RGB}{
		\includegraphics[width=0.23\linewidth]{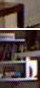}}
    \hfill
    \subcaptionbox{GT}{
		\includegraphics[width=0.23\linewidth]{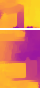}}
    \hfill
    \subcaptionbox{w/o MFA}{
		\includegraphics[width=0.23\linewidth]{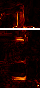}}
    \hfill
    \subcaptionbox{w/ MFA}{
		\includegraphics[width=0.23\linewidth]{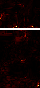}}
\end{center}
\vspace{-6mm}
    \caption{The edge part of error maps of w/ or w/o MFA on the NYUv2 dataset for $\times$4 depth super-resolution. (d) has lower prediction errors than (c), which indicates that MFA can utilize edge information in RGB better. The sizes of the cropped regions are 28$\times$40 and 58$\times$40 pixels, respectively. 
    }

\label{fig:errormap_MFA}
\end{figure}

\noindent
\textbf{Effectiveness of DDA and MFA Modules.}
We examine the effectiveness of the proposed DDA and MFA modules, and the corresponding quantitative ablated results are presented in Table \ref{tab:LADCN_GCPA_moudle}.
Meanwhile, Figure \ref{fig:errormap_DDA} and \ref{fig:errormap_MFA} shows the visualization results of error maps between the result and ground truth of with or without DDA and MFA on the NYUv2 dataset for $\times$4 upscaling, respectively.
Specifically, the removal of DDA or MFA leads to a significant performance drop in RMSE and more obvious prediction errors, which demonstrates the importance of these modules in improving the global and local precision of the reconstructed depth images. DDA and MFA can make full use of the RGB features and depth features, both modules are necessary to achieve the best performance of our proposed method.

\noindent
\textbf{Effectiveness of LDA and DGA.}
We verify the effectiveness of the learnable domain alignment block (LDA) and the dynamic geometrical alignment block (DGA) in the DDA module and replace LDA with instance normalization (IN) and batch normalization (BN) for comparison. The corresponding results are presented in Table \ref{tab:Learnable-AdaIN_DCN_moudle}.
The result reveals that independent learnable domain alignment or dynamic geometrical alignment can improve the accuracy of D2A2 significantly, and the combination of LDA and DGA yields the best result. 
Furthermore, the performance of IN or BN is worse than LDA in the presence and absence of DGA, which indicates the importance of cross-modal domain alignment.

\noindent
\textbf{Effectiveness of GC and PA.}  
We verify the effectiveness of the gated convolution (GC) and pixel attention (PA) in the MFA module and replace PA with channel attention (CA) and spatial attention (SA) for comparison. The corresponding results are presented in Table \ref{tab:GatedConv_PA_moudle}. 
It is obvious that a simple gated convolution can decrease RMSE substantially, which indicates the importance of obtaining masked features of RGB and depth and reducing interference from invalid information. 
Additionally, in comparison with the simple connection operation (baseline), CA, and SA, the PA module with baseline achieves a better SR result. It means PA has a more significant impact on cross-modal feature fusion. 
The combination of GC and PA yields the best result.

Overall, the ablation experiments demonstrate that each module in D2A2 plays a vital role in enhancing the precision of the reconstructed depth images, and the full model with both modules achieves the best performance.

\begin{table}
\caption{Quantitative results of the ablation studies for the LDA and DGA of DDA. \textbf{Bold} indicates the best score in terms of RMSE. 
}
\vspace{-6mm}	
	\begin{center}
		\begin{tabular}{c|c|c|c|c}
			\hline
			DA & DGA & $\times$4 & $\times$8 & $\times$16 \\
			\hline\hline
			$\times$ & $\times$ & 1.54 & 3.09 & 6.07\\
            $\times$ & $\checkmark$ & 1.33 & 2.68 & 5.29 \\
            IN & $\times$  &  1.97 & 3.81 & 6.41 \\
            IN & $\checkmark$  &  1.33 & 2.69 & 5.25 \\
            BN & $\times$ & 1.45 & 2.87 & 5.69\\
            BN & $\checkmark$ &  1.33 & 2.68 & 5.27 \\
			LDA & $\times$ & 1.43 & 2.86 & 5.68 \\
			LDA & $\checkmark$ & \textbf{1.32} & \textbf{2.67} & \textbf{5.19}\\
			\hline
		\end{tabular}
	\end{center}
 \vspace{-3mm}
    \label{tab:Learnable-AdaIN_DCN_moudle}
\end{table}

\begin{table}
\caption{Quantitative results of the ablation studies for the GC and PA of MFA. \textbf{Bold} indicates the best score in terms of RMSE. }
\vspace{-6mm}
\begin{center}
    \begin{tabular}{c|c|c|c|c}
        \hline
        GC & Attention & $\times$4 & $\times$8 & $\times$16 \\
        \hline\hline
        $\times$ & $\times$ & 1.54 & 3.09 & 6.07 \\
        $\checkmark$ & $\times$ &1.35 & 2.74 & 5.37\\
        $\times$ & CA & 1.95 & 3.89 & 7.55\\
        $\checkmark$ & CA & 1.94 & 3.89 & 7.58 \\
        $\times$ & SA & 1.52 & 2.99 & 5.76 \\
        $\checkmark$ & SA & 1.43 & 2.84 & 5.49 \\
        $\times$ & PA & 1.39 & 2.81 & 5.52\\
        $\checkmark$ & PA & \textbf{1.33} & \textbf{2.69} & \textbf{5.27} \\
        \hline
    \end{tabular}
\end{center}
 \vspace{-3mm}
    \label{tab:GatedConv_PA_moudle}
\end{table}
\section{Conclusion}

In this paper, we propose a Dynamic Dual Alignment and Aggregation network (D2A2) based on learnable domain alignment, dynamic geometrical alignment, gated convolution, and pixel attention. 
Our D2A2 solved the obstacles of modal and geometrical misalignment in the cross-modal RGB and depth features by dynamic dual alignment, and alleviated the challenging feature fusion problem by mask-to-pixel feature aggregation. 
Multiple indicators across various datasets demonstrate that D2A2 is the state of the art for guided depth super-resolution. 
We are optimistic that our innovative approach to alignment and feature fusion will serve as a source of inspiration for future GDSR tasks.

{
    \small
    \bibliographystyle{ieeenat_fullname}
    \bibliography{main}
}


\end{document}